\documentclass[letterpaper, 10 pt,conference]{ieeeconf}

\usepackage{graphicx}
\usepackage{setspace}
\usepackage{graphicx}
\usepackage{caption}
\usepackage{subcaption}
\usepackage{amsmath}
\usepackage{amssymb}
\usepackage{algorithm}
\usepackage{algorithmic}
\usepackage{xcolor}
\usepackage{booktabs}
\usepackage{multirow}
\usepackage[table]{colortbl}
\usepackage{diagbox}
\usepackage{makecell}
\usepackage[
    backend=biber, 
    bibencoding=utf8,
    sorting=none,
    firstinits=true, doi=false,isbn=false,url=false,eprint=false,
    style=ieee,
    hyperref, 
]{biblatex} 

\let\labelindent\relax
\usepackage{enumitem}
\usepackage{tikz}
\usepackage[nolist]{acronym}
\usepackage[most]{tcolorbox}

\usepackage[final]{changes}
\tcbuselibrary{listings, breakable, skins}

\begin{acronym}[MPC]
\acrodef{ACF}{Aggregated Channel Features}
\acro{AIP}{Anterior Parietal Lobe}
\acro{ANN}{Artificial Neural Networks}
\acro{BO}{Bayesian Optimization}
\acro{CAD}{Computer-Aided Design}
\acro{CIP}{Caudal Intraparietal Sulcus}
\acro{CNN}{Convolutional Neural Network}
\acro{DNN}{Deep Neural Networks}
\acro{FOV}{Field of View}
\acro{DCNN}{deep convolutional neural network}
\acro{GMM}{Gaussian Mixture Model}
\acrodefplural{GMM}{Gaussian Mixture Models}
\acro{GP}{Gaussian Proccess}
\acrodefplural{GP}{Gaussian Proccesses}
\acro{GPU}{Graphical Proccessing Unit}
\acrodefplural{GPU}{Graphical Proccessing Units}
\acro{GHT}{Generalized Hough Transform}
\acro{HOG}{Histogram of Gradients}
\acro{HRI}{Human Robot Interaction}
\acro{HVS}{Human Visual System}
\acro{IOR}{Inhibition of Return}
\acro{KL}{Kullback-Leibler}
\acro{JPDA}{Joint Probabilistic Data Association}
\acro{LGN}{Lateral Geniculate Nucleus}
\acrodefplural{LSTM}[LSTM]{Long Short-Term Memory Units}
\acro{MAB}{Multi-Armed Bandit}
\acro{MAP}{Maximum a Posteriori}
\acro{MCTS}{Monte Carlo Tree Search}
\acro{MOT}{Multiple Object Tracking}
\acro{NBV}{next-best-view}
\acro{POMDP}{Partially Observable Markov Decision Process}
\acrodefplural{POMDP}{Partially Observable Markov Decision Processes}
\acrodefplural{GMM}[GMM]{Gaussian Mixture Models}
\acro{RANSAC}{Random Sample Consensus}
\acro{RF}{Receptive Field}
\acrodefplural{RF}{Receptive Fields}
\acro{RNN}{Recurrent Neural Network}
\acro{RPN}{Region Proposal Network}
\acro{SIFT}{Scale Invariant Feature Transform}
\acrodef{SVM}{Support Vector Machine}
\acrodefplural{SVM}{Support Vector Machines}
\acro{UCB}{Upper Confidence Bound}
\acrodefplural{UCB}{Upper Confidence Bounds}
\acro{UT}{Unscented Transform}
\acro{WTA}{Winner Take All}
\acro{RoI}{Region of Interest}
\acro{FPS}{frames per second}
\acro{MOTP}{Multiple Object Tracking Precision}
\acro{SLAM}{simultaneous localization and mapping}
\acro{RRT}{rapidly-exploring random tree}
\acroplural{RRT}[RRTs]{rapidly-exploring random trees}
\acro{UAV}{unmanned aerial vehicle}
\acroplural{UAV}[UAVs]{unmanned aerial vehicles}
\acro{IMU}{inertial measurement unit}
\acroplural{IMU}[IMUs]{inertial motion units}
\acro{VIO}{Visual Inertial Odometry}
\end{acronym}
\IEEEoverridecommandlockouts
\pdfoutput=1
\begin{document}

\title{Real-Time Volumetric-Semantic Exploration and Mapping: An Uncertainty-Aware Approach}

\author{Rui Pimentel de Figueiredo, Jonas le Fevre Sejersen, Jakob Grimm Hansen,\\ Martim Brand\~ao and Erdal Kayacan
\thanks{R. Figueiredo, J. Sejersen, J. Hansen, E. Kayacan are with Artificial Intelligence in Robotics Laboratory (AiR Lab), the Department of Electrical and Computer Engineering, Aarhus University, 8000 Aarhus C, Denmark
        {\tt\small \{jonas.le.fevre,rui,erdal\} at ece.au.dk}
M. Brand\~ao is with King's College London (KCL), London, UK
        {\tt\small martim.brandao at kcl.ac.uk}}
}
\maketitle
\begin{abstract}
In this work we propose a holistic framework for autonomous aerial inspection tasks, using semantically-aware, yet, computationally efficient planning and mapping algorithms. The system leverages state-of-the-art receding horizon exploration techniques for \ac{NBV} planning with geometric and semantic segmentation information provided by state-of-the-art \acp{DCNN}, with the goal of enriching environment representations. The contributions of this article are threefold, first we propose an efficient sensor observation model, and a reward function that encodes the expected information gains from the observations taken from specific view points. Second, we extend the reward function to incorporate not only geometric but also semantic probabilistic information, provided by a \ac{DCNN} for semantic segmentation that operates in real-time. The incorporation of semantic information in the environment representation allows biasing exploration towards specific objects, while ignoring task-irrelevant ones during planning. Finally, we employ our approaches in an autonomous drone shipyard inspection task. A set of simulations in realistic scenarios demonstrate the efficacy and efficiency of the proposed framework when compared with the state-of-the-art.
\end{abstract}

\IEEEpeerreviewmaketitle

\section{Introduction}
\label{sec:intro}
This paper is focused on developing autonomous navigation, and mapping solutions for vision-based exploration and mapping of shipyard scenarios using \acp{UAV}. Inspection of a single vessel is costly, time consuming, and demanding as it requires manually categorizing damages and paint deterioration from low quality pictures. It is also risky since it has to be performed by human climbers through the use of scaffolding's and mobile cranes. The use of \acp{UAV} offers a more cost effective and safer solution, since it avoids the need of direct intervention by human operators. 
\begin{figure}[t]
  \centering
  \begin{subfigure}{0.97\textwidth}
     \includegraphics[width=0.5\textwidth]{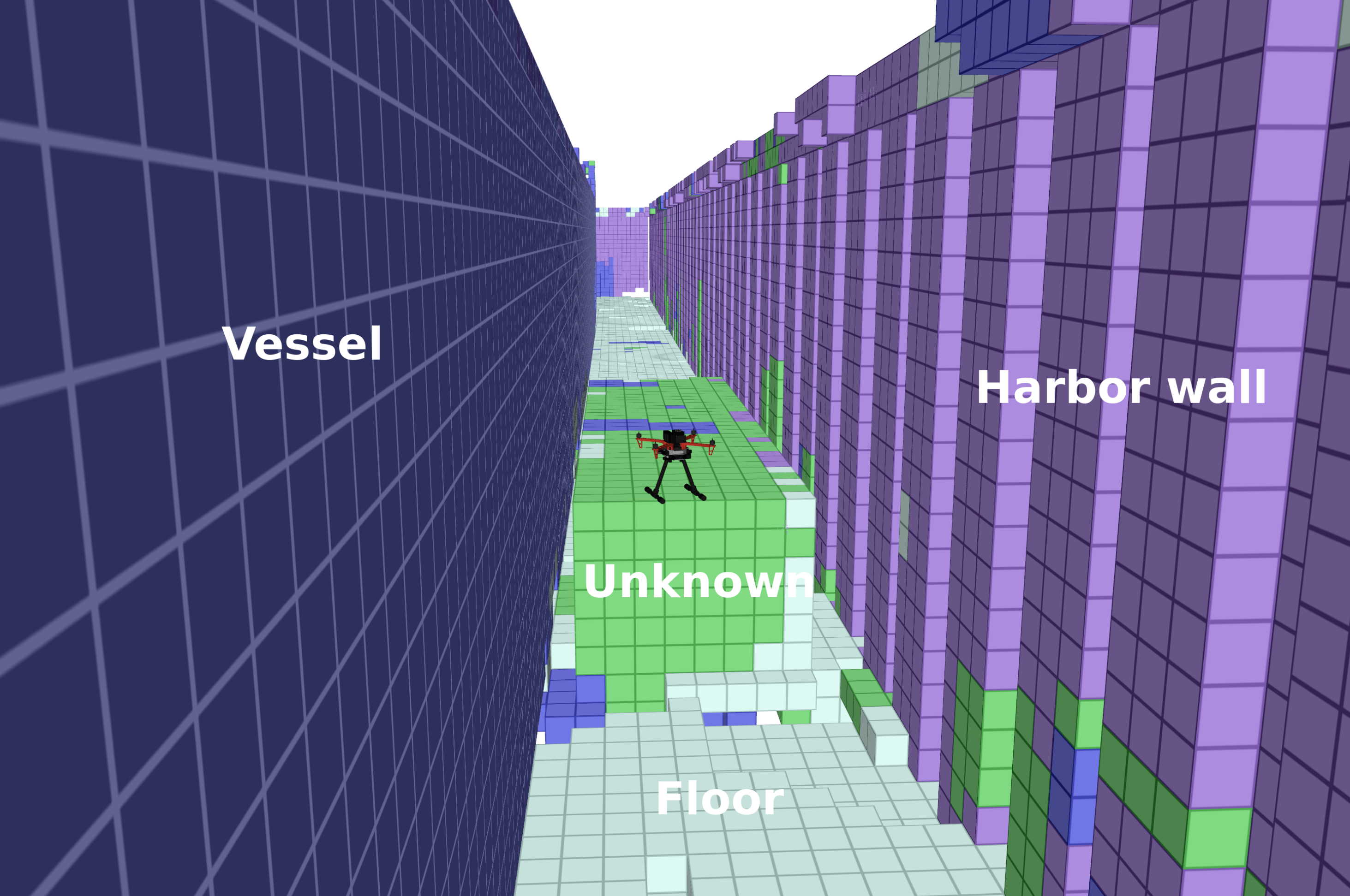}  
  \end{subfigure}
\caption{A volumetric-semantic mapping representation of a simulated live-mission of an autonomous drone shipyard inspection task obtained using our volumetric and semantically uncertainty-aware receding horizon path planner.}  
  \label{fig:RSUencountered} 
\end{figure}
We propose a solution for autonomous exploration and mapping using a single \ac{UAV}, that combines probabilistic semantic-metric probabilistic measures fused in an efficient octogrid map representation, that are used by a fast geometric and semantically aware \ac{NBV} planning algorithm, to efficiently reconstruct and label all objects.


The main contributions of this work are as follows:
\begin{enumerate}
    \item An efficient probabilistic observation model for depth and semantic information 
    provided by consumer grade RGB-D sensors and state-of-the-art \acp{DCNN} for 2{D} scene semantic segmentation. 
    \item A probabilistic mapping formulation that allows Bayesian fusion of volumetric and semantic information in a memory-efficient 3{D} octogrid structure.
    \item A real-time, receding-horizon probabilistic path planning approach, that considers both geometric and semantic probabilistic information for planning, using \acp{RRT}. Our method is flexible and allows biasing exploration towards specific object classes.
    \item A thorough evaluation of the advantages of our semantically-guided mapping approach when compared with the state-of-the-art in a realistic simulation environment.
\end{enumerate}

The rest of this article is organized as follows: first, in Section \ref{sec:related_work}, we overview in detail the literature on semantic and geometric environment representations as well as \ac{NBV} planning algorithms for autonomous mapping and robotics applications. Then, in Section \ref{sec:methodology} we describe the proposed methodologies for autonomous drone inspection tasks. The results Section \ref{sec:results} evaluates the proposed strategies for autonomous mapping and navigation, in a realistic simulation shipyard environment. Finally, in Section \ref{sec:conclusions}, some conclusions are drawn from this study and suggestions for future research are proposed.

\section{Related Work} 
\label{sec:related_work}
In this section we overview the state-of-the-art on environment representations, and active perception algorithms for autonomous localization and mapping, used in robotics applications.
\subsection{Map representations} 

The most common metric mapping representation in the literature is called probabilistic 3D occupancy grids. Such maps represent the environment as a block of cells, each one having a binary state (either occupied, or free). They are popular in the robotics community since they simplify collision checking and path planning, access is fast and memory use can be made efficient through octrees~\cite{hornung2013octomap}.

Semantic representations typically extend metric representations such as images and grids with a semantic layer, and rely on segmentation methods.
Methods for 2{D} image semantic segmentation deal with the task of associating each pixel of an image as belonging to one of a set of different known classes. 
State-of-the-art methods for image segmentation are based on deep neural network architectures \cite{ronneberger2015u,He_2017_ICCV} that learn from large annotated datasets to regress from 2{D} images to object masks that encode the layout of input object’s. 
\deleted{One of the issues with modern approaches is that they usually compromise spatial resolution to achieve real-time inference speed, which leads to poor accuracy performance.} The Bilateral Segmentation Network (BiSeNet) \cite{bisenet2018}, offers a way of balancing effectively the trade-off between speed and accuracy. BiSeNet \cite{bisenet2018} first comprises a spatial path with a small stride to preserve spatial information and generate high-resolution feature maps, and a context path with a fast down-sampling mechanisms for the generation of high resolution receptive fields. These are efficiently merged, using a feature fusion module.
In this work we use BiseNet for image semantic-segmentation, because it is compact, fast, robust \cite{hu2019comparison}, and easy to use, making in suitable for remote sensing applications running on embedded systems (e.g. UAVs) with low budget computational specifications. 

\subsection{Active perception}

In this paper we tackle the problem of actively controlling the viewpoint(s) of a sensor(s) to improve task performance, which is of the utmost importance in robotics applications \cite{aloimonos1988active}. 

\subsubsection{Next-best-view planning}
\ac{NBV} planning has been widely studied by the robotics community and  plays a role of primordial importance on object reconstruction \cite{volNBVManufacturing2019}, autonomous mapping \cite{isler2016information}, and safe navigation \cite{brandao2020placing} tasks, to name a few.
Existing \ac{NBV} approaches~\cite{delmerico2018comparison} belong to one of two main categories: frontier-based and information-driven planning. Frontier-based planners \cite{frontier_exploration} guide the robot to boundaries between unknown and free space, which implicitly promotes exploration. Information-driven methods rely on probabilistic environment representations, and select the views that maximize expected information gains~\cite{delmerico2018comparison} by back-projecting probabilistic volumetric information on candidate views via ray casting. \deleted{Methods in the literature differ in the way they define information gain.} One approach to the problem is to incrementally compute and target a sensor at the \acp{NBV} according to some criteria (e.g. maximize 3D reconstruction quality). \deleted{For example, \cite{Brandao2013isrm} proposes a simple \ac{NBV} algorithm which greedily targets the gaze of a humanoid robot at points of maximum entropy along a robot trajectory.}
Instead of just considering the entropy,~\cite{isler2016information} proposes a set of extensions to \cite{kriegel2015efficient}'s information gain definition, including the incorporation of visibility probability as well as the likelihood of seeing new parts of the object \cite{brandao2020placing}. 
Incremental sampling techniques (i.e. random tree sampling)~\cite{lavalle1998rapidly,rapid_random_trees_star2011} sample the view space in a tree manner, using \acp{RRT} methods. 
Since taking all the possible views into consideration is computationally intractable, RRT-based approaches consider only a subset of possible views at each planning step. The tree is randomly expanded throughout the exploration space, and each branch forms a group of random subbranches. 
The quality of each branch is determined according to some criteria (e.g. by the amount of unmapped space that can be explored \cite{bircher2016receding,bircher2018receding}), and a receding-horizon approach is used at each planning step, until complete exploration of the environment is achieved.
\deleted{Recent, more modern approaches are based on deep learning techniques and bypass the need of explicit traditional sensing-planning-acting architectures. For instance, in the work of \cite{deepReinforcementLearning2020} the authors propose an end-to-end motion planner for UAVs. Their method creates desirable motion plans using raw depth images and correlations between local spatial portions of images to generate desirable motion primitive sequences on the fly, which can be learned using reinforcement learning techniques \cite{kayacan2019swift}.
In \cite{Gupta_2017_CVPR} the authors propose a neural architecture called Cognitive Mapper and Planner (CMP) for planning and mapping in unknown environments. The proposed architecture learns to map from first-person views and plans a sequence of actions towards semantically specified goals in the environment. CMP constructs a belief map of the world and applies a neural net planner to produce the next action at each time step.} 

\section{Methodologies}
\label{sec:methodology}
In the rest of this section we describe the proposed system and methodologies for active exploration and semantic-metric mapping of man-made infrastructures.

\subsection{System overview}

The proposed drone system for autonomous inspection consists of a forward-facing camera, \ac{IMU}, and an altimeter. Our navigation system relies on an off-the-shelf \ac{SLAM} system with loop closing and relocalization capabilities \cite{orb_slam2_2017}, which is fed with RGB-D and \ac{IMU} measurements, for improved robustness on self-motion tracking performance \cite{rovio2015}. 

\deleted{We propose a novel probabilistic observation model that combines metric and semantic visual cues, which are efficiently fused in a volumetric octogrid structure \cite{hornung2013octomap}, and a \ac{NBV} planner that leverages both geometric and semantic information provided by RGB-D data, for task-dependent exploration.}



\subsection{Probabilistic volumetric-semantic occupancy mapping}
We consider an octomap representation, defined as a 3D uniform voxel grid structure that encloses the workspace around the robot, represented by $m=\{m_i\}$, where each voxel $m_i=\{m^o_i, m^s_i\}$ with $m^o_i\in \{0,1\}$ being a binary random variable representing its occupancy, and $m^s_i\in \{1,...,K_c\}$ a semantic variable representing its object class. We use recursive Bayesian volumetric mapping~\cite{thrun2005probabilistic} to sequentially estimate the posterior probability distribution over the map, given sensor measurements $z_{1:t}=\{z^{o}_{1:t};z^{s}_{1:t}\}$ and sensor poses $p_{1:t}$ obtained through the robot kinematics model and an off-the-shelf \ac{SLAM} module, from time $1$ to $t$ 
\begin{equation}\label{eq:map_posterior}
P(m|z_{1:t},p_{1:t})=\prod_{i} P(m_i|z_{1:t},p_{1:t})
\end{equation}
assuming the occupancy of individual cells are independent.
Filtering updates can be recursively computed in log-odds space~\cite{moravec1985high} to ensure numerical stability and efficiency.

\subsection{3{D} semantic segmentation}
\begin{figure*}[t!]
    \vspace{0.2cm}
    \begin{subfigure}[b]{0.98\textwidth}
        \includegraphics[width=\textwidth]{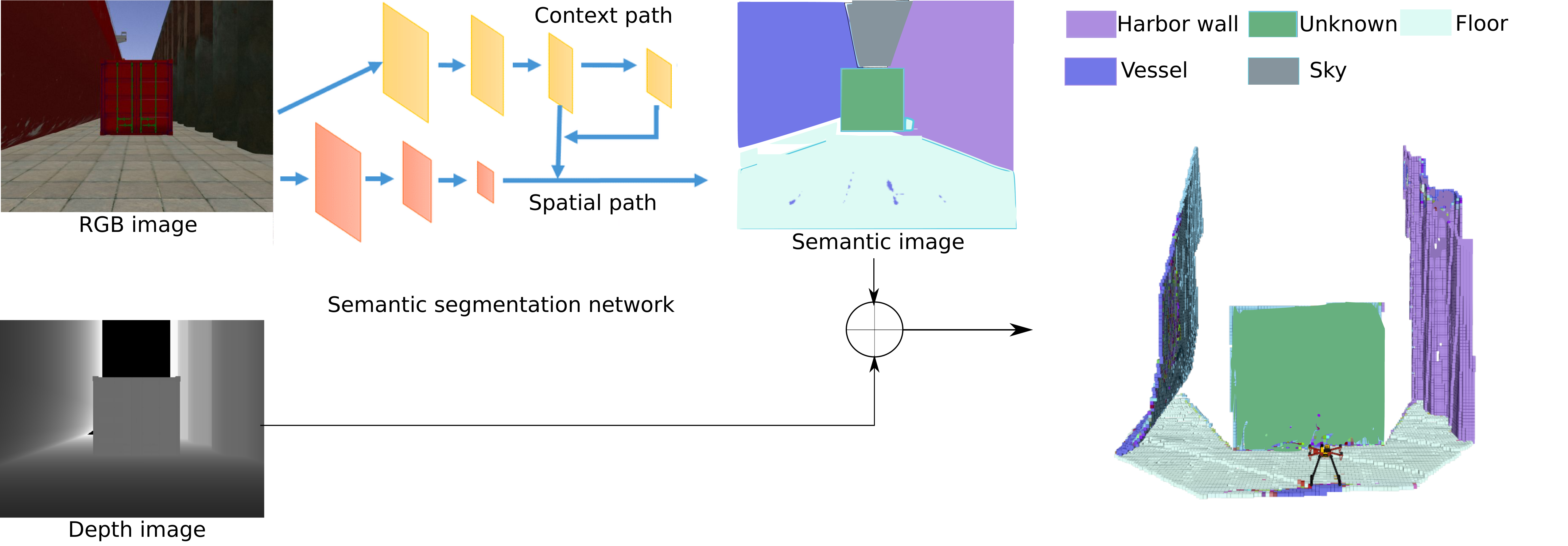}
        \caption{3D semantic segmentation system architecture}
        \label{fig:semantic_pcl}
    \end{subfigure}
    \centering
    \begin{subfigure}[b]{0.49\textwidth}
        \includegraphics[width=\textwidth]{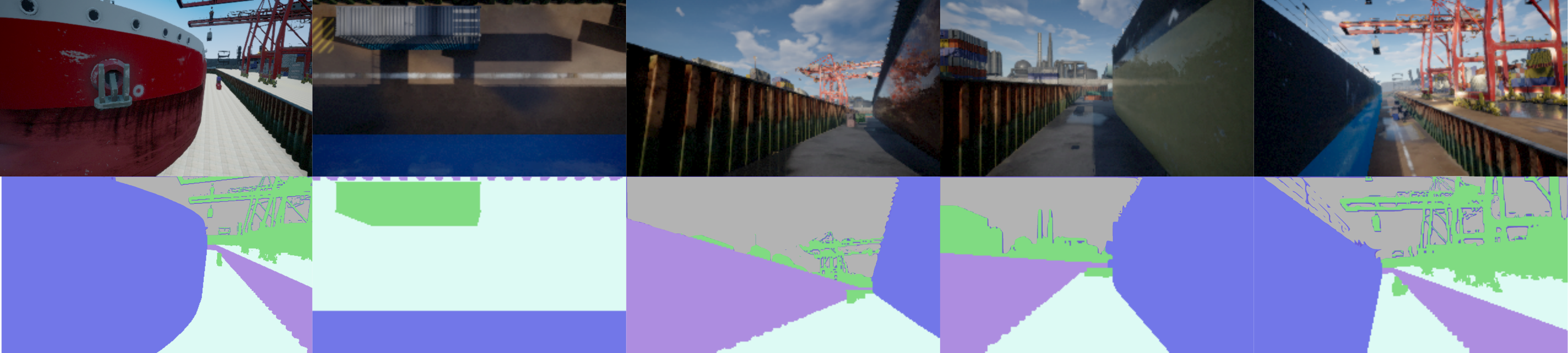}
        \caption{AirSim}
        \label{fig:airsim_samples}
    \end{subfigure}
    \begin{subfigure}[b]{0.49\textwidth}
        \includegraphics[width=\textwidth]{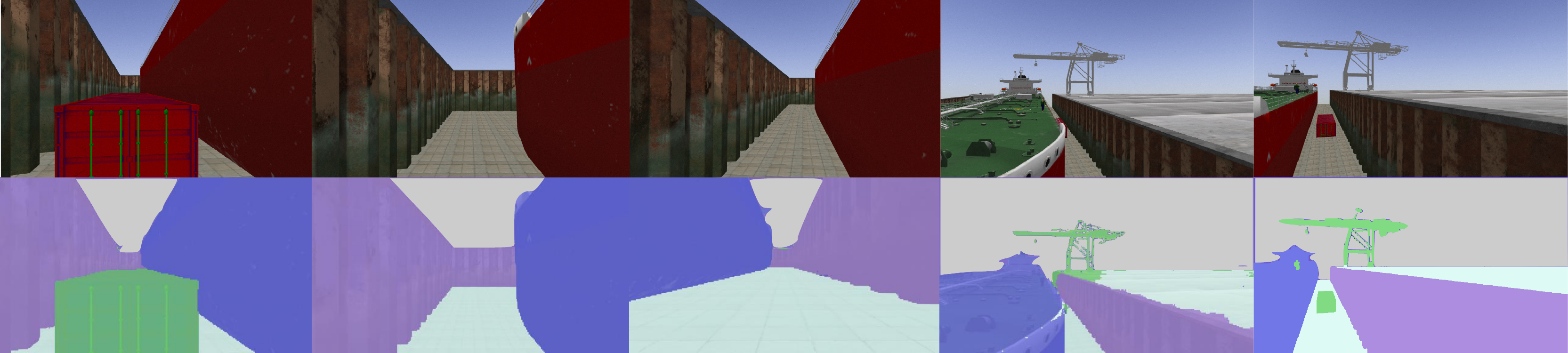}
        \caption{Gazebo}
        \label{fig:gazebo_samples}
    \end{subfigure}
\caption{(a) Proposed architecture for 3D semantic segmentation. (b) and (c) Top: sample training (b) and testing (c) images collected in Unreal Engine and Gazebo, respectively. Bottom: labelled (b) and inferred'(c) images including ship (violet), sky (grey), floor (light blue), harbor walls (purple), and unknown harbor objects such as containers and cranes (green).}
\label{fig:semantic_segmentation_architecture}
\end{figure*}

Our method for semantic segmentation relies on a \ac{DCNN} encoder-decoder segmentation network, named BiseNet \cite{bisenet2018}. BiseNet receives RGB or grayscale images as input, encodes the image information, then decodes it again, and outputs a probability distribution over the known object categories for each pixel $(u,v)$. \deleted{In this work we use BiseNet because it is compact, fast, robust and easy to use, being suitable for remote sensing applications running on embedded systems (e.g. UAVs) with low computational specifications \cite{hu2019comparison}.}

To obtain the 3{D} representation in real-time, we fuse the 2{D} semantic information provided by BiseNet with depth measurements provided by consumer grade RGB-D cameras. 
For each pixel $(u,v)$ belonging to an input image $I$, the network outputs a probability distribution $p^c(u,v)\in\mathcal{P}^{K_c}$ over the set of known classes $\mathcal{C}$, where $K_c$ represents the number of known classes. For training the network we use the categorical Cross-Entropy loss function
\begin{align}
    CE = -\text{log}\left(\frac{e^{s_p}}{\sum_{j}^C e^{s_j} }\right)
\end{align}
where $s_j$ is the \ac{DCNN} output score for the class $j\in \mathcal{C}$, and $s_p$ the positive class. 

At run-time, at each time instant $t$, the probability distributions over all classes and image pixels (i.e. probabilistic semantic image), obtained with BiseNet, are merged with the corresponding depth image. We rely on a known extrinsic calibration parametric model, to obtain a semantic probabilistic point cloud (see Fig. \ref{fig:semantic_pcl}), where each point has a semantic probability distribution given by
\begin{align}
    P(z^s_{t,k}|I_t,p_t)
\end{align}

\subsubsection{Semantic dataset}
Supervised training of deep neural networks relies on the availability of large annotated data sets, hand-labeled in a laborious and time consuming manner, which may be impracticable for applications and machine learning tools, requiring large data sets. We train our BiseNet model with synthetic data, generated in a realistic simulation environment to overcome the reality gap.
The dataset used for training our semantic segmentation network is generated using a combination of Unreal Engine 4 to create a realistic dock environment and AirSim \cite{shah2018airsim} to extract the images and the corresponding labelled images, gathered by a multi-camera UAV (see Fig. \ref{fig:semantic_segmentation_architecture}). 
\begin{table*}[t]
\centering
\normalsize
\vspace{0.2cm}
\caption{Dataset used for training (AirSim), validating (AirSim) and testing (Gazebo) the proposed scene semantic segmentation network, for shipyard environments. The dataset specifications include the number of images and classes in each partition.}
\begin{tabular}{|l|c|c|c|c|c|c|c|}
\hline
                & \# of Images & \# Sky  & \# Floor & \# Ship       & \# Harbor wall    & \# Unknown \\ \hline
Training data   & 49648        & 35353   & 45833    & 35281         & 28406             & 34951    \\ \hline
Validation data & 9930         & 7088    & 9177     & 7027          & 5587              & 6953    \\ \hline
Test data       & 184          & 175     & 174      & 153           & 177               & 72      \\ \hline
\end{tabular}
\label{tab:dataset_stats}
\end{table*}
\deleted{The UAV used for recording the dataset multiple cameras, four mounted on each side of the drone and one on the bottom. These cameras generate labeled images while the UAV is performing a flying route (see Fig. \ref{fig:semantic_segmentation_architecture}). The environment is a dry harbor containing multiple ships with different textures, colored ship containers, two cranes, and tiled/concrete floor, as well as multiple objects (e.g. puddles of water) scattered around on the floor.} In order to increase the variability of the acquired dataset, shadows and illumination conditions are dynamically changed based on the simulated time of the day and weather conditions. For quantitatively testing the performance of our semantic segmentation networks, we have generated a dataset in Gazebo in a shipyard environment (see Table \ref{tab:dataset_stats}).

\subsection{Efficient probabilistic sensor fusion model}
\label{sec:data_fusion}

Our sensor depth noise model assumes that single point measurements $z^{o}_{t,k}$ are normally, independent and identically distributed (iid) according to

\begin{equation}
z^{o}_{t,k}|m,p_{t} \sim \mathcal{N}(z^{o*}_{t,k};\Sigma^{o}_{t,k})
\end{equation}
with
\[
\Sigma^{o}_{t,k}=\text{diag}(\sigma^l_{t,k},\sigma^l_{t,k},\sigma^a_{t,k})
\]
where $z^{o*}_{t,k}$ denotes the true location of the measurement and $\sigma^l_{t,k}$ and $\sigma^a_{t,k}$ represent the lateral and axial noise standard deviations, respectively.
We assume that noise is predominant in the axial direction ($\sigma^a>>\sigma^l$) which allows approximating the {3D} covariance matrix by a {1D} variance. 
For each measurement $z^{o}_{t,k}$, we then update the corresponding closest grid cell $m_i$ as follows
\begin{equation}
\label{eq:sensor_fusion}
P(m^o_{i,t}|z^{o}_{t,k},p_{t})\approx
F_{z}\left(z^{o}_{t,k}+\frac{\delta}{2}\right)-F_{z}\left(z^{o}_{t,k}-\frac{\delta}{2}\right)
\end{equation}
where $\delta$ represents the grid resolution and $F_{z}(.)$ the cumulative normal distribution function of $z$,
where the axial error standard deviation can be approximated by a simple quadratic model
\begin{equation}
\sigma^a_{t,k}\approx\lambda_a \|z^o_{t,k}\|^2
\end{equation}
with $\lambda_a$ being a sensor specific scaling factor. All other cells belonging to the set of voxels traversed through ray casting (from the origin to end point $z^o_i$) are updated as being free with probability $P_\text{free}$. This is a reasonable approximation, considering that map resolution and sensory noise have the same order, while at the same time allowing to significantly reduce the amount of computation.
The associated semantic measurements $z^s_{t,k}$, with probabilities given by $P ( z^s_{t,k} |I_t, p_{t,k} )=\{P_1,...,P_{K_c}\}$  are updated independently, using
\begin{align}
P(m^{s}_{i,t}|z^{s}_{t,k},p_{t})=\eta P ( z^s_{t,k} |I_t, p_{t} )P(m^{s}_{i,t-1}|z^{s}_{1:t-1},p_{1:t-1})
\end{align}
where $\eta$ is a normalizing constant.

\subsection{Semantic-aware next-best-view planning}

The proposed receding horizon \ac{NBV} planner is based on the one of \cite{bircher2018receding}.  At each viewpoint, the planner generates a set of rays $R$ that end if they collide against a physical surface or reach the limit of the map.
For a given occupancy map representing the world $m$, the set of visible and unmapped voxels from configuration $\xi_k$ is represented by $\text{V}(m,\xi_k)$.  Every voxel $m_i \in \text{V}(m,\xi_k)$ lies in  the unmapped  exploration  area and is visible by the sensor in configuration $\xi_k$ (i.e. unoccluded and within the field of view).
The expected information gain $\text{G}(n_k)$ for a given tree node $n_k$ is the cumulative information gain collected from all nodes $n_1,...,n_k$, i.e. all nodes along the path from $n_1$ (the robot's current pose and RRT root) to $n_k$. This gain function is defined as
\begin{align}
\text{G}(n_k)=\text{G}(n_{k-1})+
\left( \sum_{m_i \in \text{V}(m,\xi_k)} I_g(m_i) \right) e^{-\lambda\text{cost}\left(\sigma^k_{k-1}\right)}
\end{align}
where $n_k$ is node $k$ in the \ac{RRT}, and the exponential is a weighting factor to favor low distance paths.

The original formulation of \cite{bircher2018receding} models per-voxel information gain $I_g$ as
\begin{equation}
I_g(m_i)=\begin{cases}
    1\quad    & \text{if } P(m^o_i)	= 0.5 (\text{i.e. unknown})  \\
    0\quad    & \text{otherwise (i.e. free or occupied)}  \\
\end{cases} 
\end{equation}
where $P(m^o_i)$ is the probability of cell $i$ being occupied, i.e. they consider only volumetric information.
Our method differs in the way information gain is defined. It leverages both the volumetric and semantic information entailed by each voxel. We model volumetric entropy at each voxel $m_i$ as
\begin{align}
H^o(m_i) = -P(m^o_i) ln( P(m^o_i) ) 
\end{align}
and the semantic entropy as a sum over per-class entropy 
\begin{equation}
H^s(m_i) = \sum^{K_C}_{k=1} H^s_k(m_i) 
\end{equation}
with per class-entropy $H^s_k(m_i)$ equal to
\begin{equation}
H^s_k(m_i) = - P(m^{s_k}_i)  ln( P(m^{s_k}_i) )
\end{equation}
where $P(m^{s_k}_i)$ is the probability of cell $i$ being of class $k$.

We propose two different probabilistic information gain formulations. The first formulation is agnostic to semantics and accounts only for the occupancy information the voxel provides, according to
\begin{equation}
    I_g(m_i) = -H^o(m_i)
\end{equation}
The second formulation incorporates semantic constraints as a weighted summation of the information gain per class, across all voxels:
\begin{equation}
    I_g(m_i) = H^o(m_i)\sum_k \mathbf{w}^{s}_k H^{s}_k(m_i) { \quad } \sum_k{\mathbf{w}^{s}_k = 1}
\end{equation}
where $\mathbf{w}^s=\left[{w}^{s}_{1},...,{w}^{s}_{K_c}\right]$ corresponds to a user specified weight vector, representing task-dependent class specific exploration biases. 
As in \cite{bircher2018receding}, the path which gets executed by the robot is picked by choosing the largest-gain node in the RRT, and executing the first segment of the path towards that node in a receding-horizon manner.




\section{Experiments}

\label{sec:results}
In this section we conduct a set of experiments in a realistic simulation environment to assess the performance of the proposed holistic approach in a drone-aided shipyard inspection scenario. We use a quadcopter provided with a front-side RGB-D camera and an \ac{IMU}. Our first goal was to assess the performance of the proposed pipeline on autonomous mapping the shipyard environment. 
All experiments were run on an Intel\textregistered~i7-10875H CPU with a GeForce RTX 2080 graphics card. 

\subsection{Semantic segmentation}
\label{sec:experiments}
\paragraph{Semantic Segmentation Evaluation Metrics}
In order to access the performance of the image segmentation module, we rely on the pixel accuracy $P_\text{acc}(C)$ metric:
\begin{align}
    P_\text{acc}(c) =\frac{\#TP+\#TN}{\#TP+\#TN+\#FP+\#FN}
\end{align}
where true positive (TP), false positive (FP), true negative (TN) and false negatives (FN), represent pixels classified correctly as $c$, incorrectly as $c$, correctly as not $c$ and incorrectly as not $c$, respectively. 

\begin{figure}[h]
	\centering
	\includegraphics[width=0.49\textwidth]{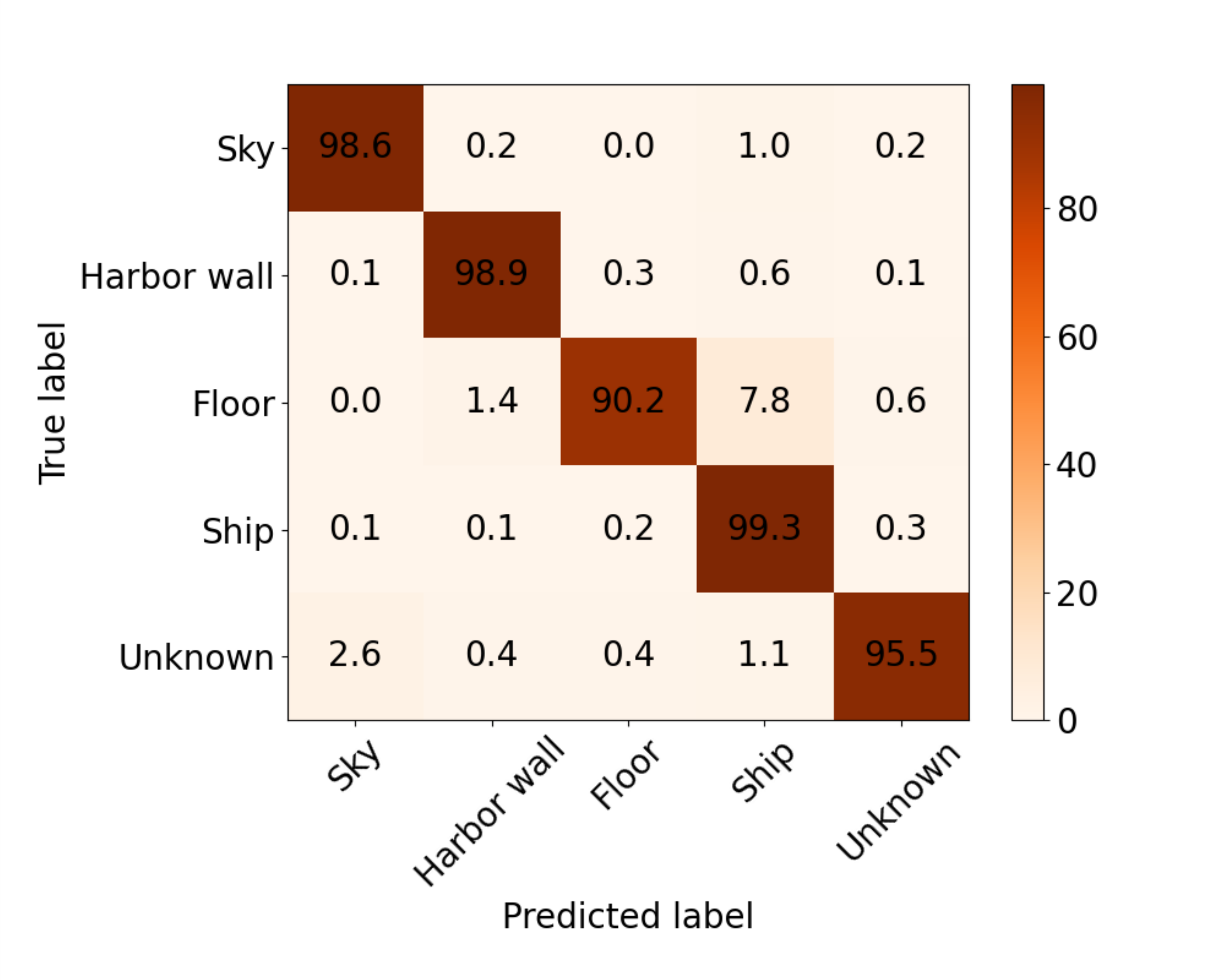}
	\caption{Pixel confusion matrix for the semantic network, for 93\% IoU score}
	\label{fig:confusion_matrix}
\end{figure}

We have quantitatively assessed the performance of our BiseNet network model \cite{bisenet2018}) for an input size of $512\times288$ using the dataset described in Table \ref{tab:dataset_stats}. Figure \ref{fig:confusion_matrix} shows the resulting confusion matrix on the test set (Gazebo), and demonstrates the network capability of learning to correctly classify different structures in the shipyard environment, with an overall accuracy of $97.7\%$ on the test set (see Table \ref{tab:table_results}), without the need of domain randomization and adaptation techniques to bridge the gap between the two different domains. 

\begin{table}[h!]
\centering
\caption{Semantic segmentation network performance on the validation (AirSim) and test set (Gazebo).} 
\begin{tabular}{|l|l|l|l|}
\hline
             &Overall Acc & Mean Acc & Mean IoU      \\ \hline
Val          & 0.944      & 0.959    & 0.922    \\ \hline
Test         & 0.977      & 0.965    & 0.930    \\ \hline
\end{tabular}
\label{tab:table_results}
\end{table}

\subsection{Planning system evaluation}

\subsubsection{Shipyard simulation scenario}
\begin{figure}[t]
    \vspace{0.2cm}
	\includegraphics[width=0.24\textwidth]{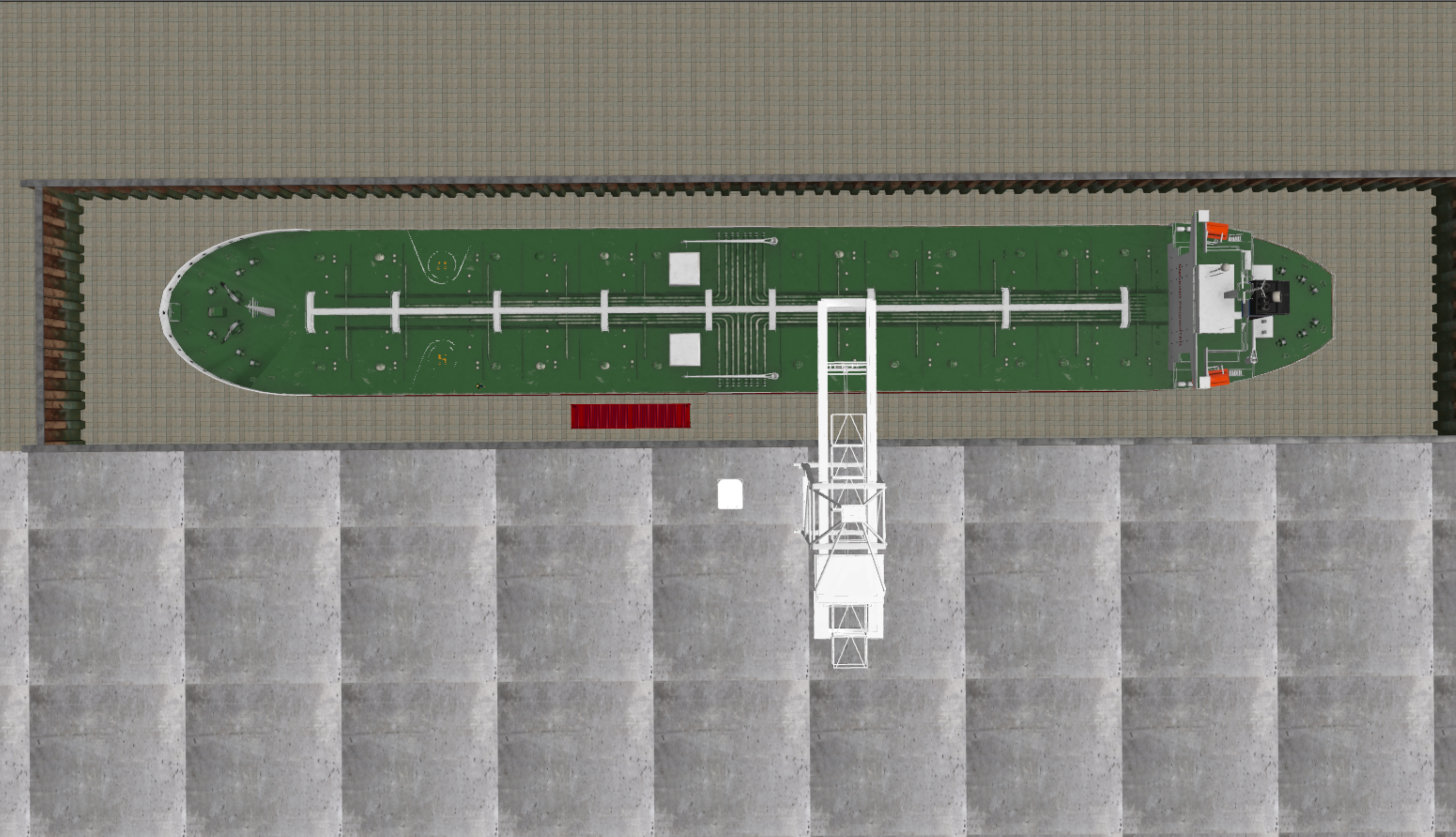}
	\includegraphics[width=0.24\textwidth]{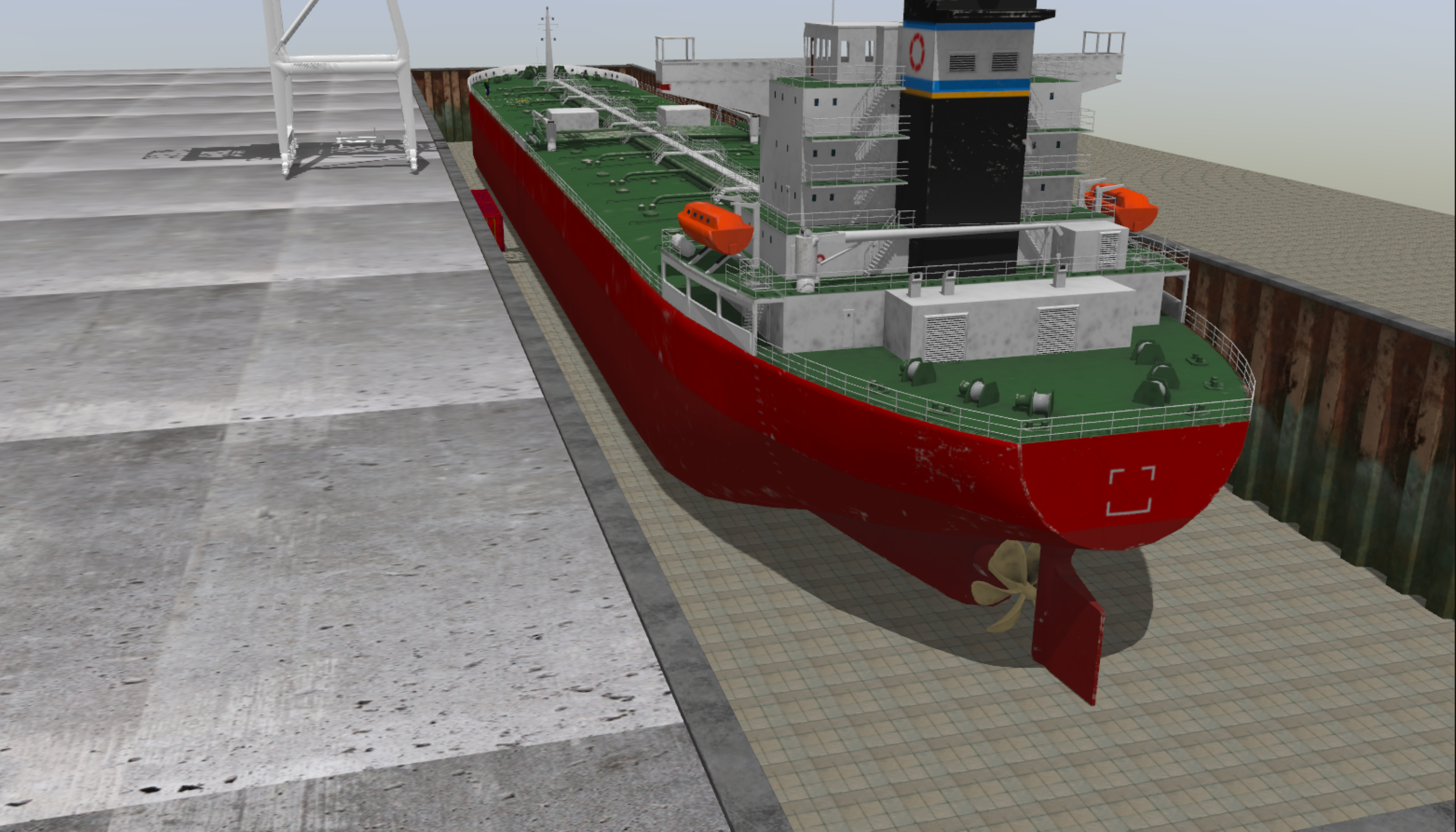}
	\caption{Scenario used for evaluation in (Gazebo/ROS) simulation studies.}
	\label{fig:simulation_scenario}
\end{figure}

In order to be able to quantitatively and qualitatively measure the performance of the proposed mapping and planning approaches, a realistic shipyard environment (see Fig. \ref{fig:simulation_scenario}) was created using the Gazebo simulator \cite{gazeboSimulationEnvironment}. The environment consists of a dry-dock measuring $145\times 30\times 8 m$, containing a crane, a container and a dry dock. An intelligent active mapping algorithm should maximize task-related rewards, in this case information gathering, by focusing on rewarding viewing directions. We constrain the planner to output positions within this volume, since mapping and inspecting the top of the UAV is time-consuming and expensive (i.e.  every  hour  in  the  dry  dock  is extremely costly).

\begin{figure*}[t]
	\centering
	\begin{subfigure}[b]{1.0\textwidth}
		\centering
		\includegraphics[width=0.23\textwidth]{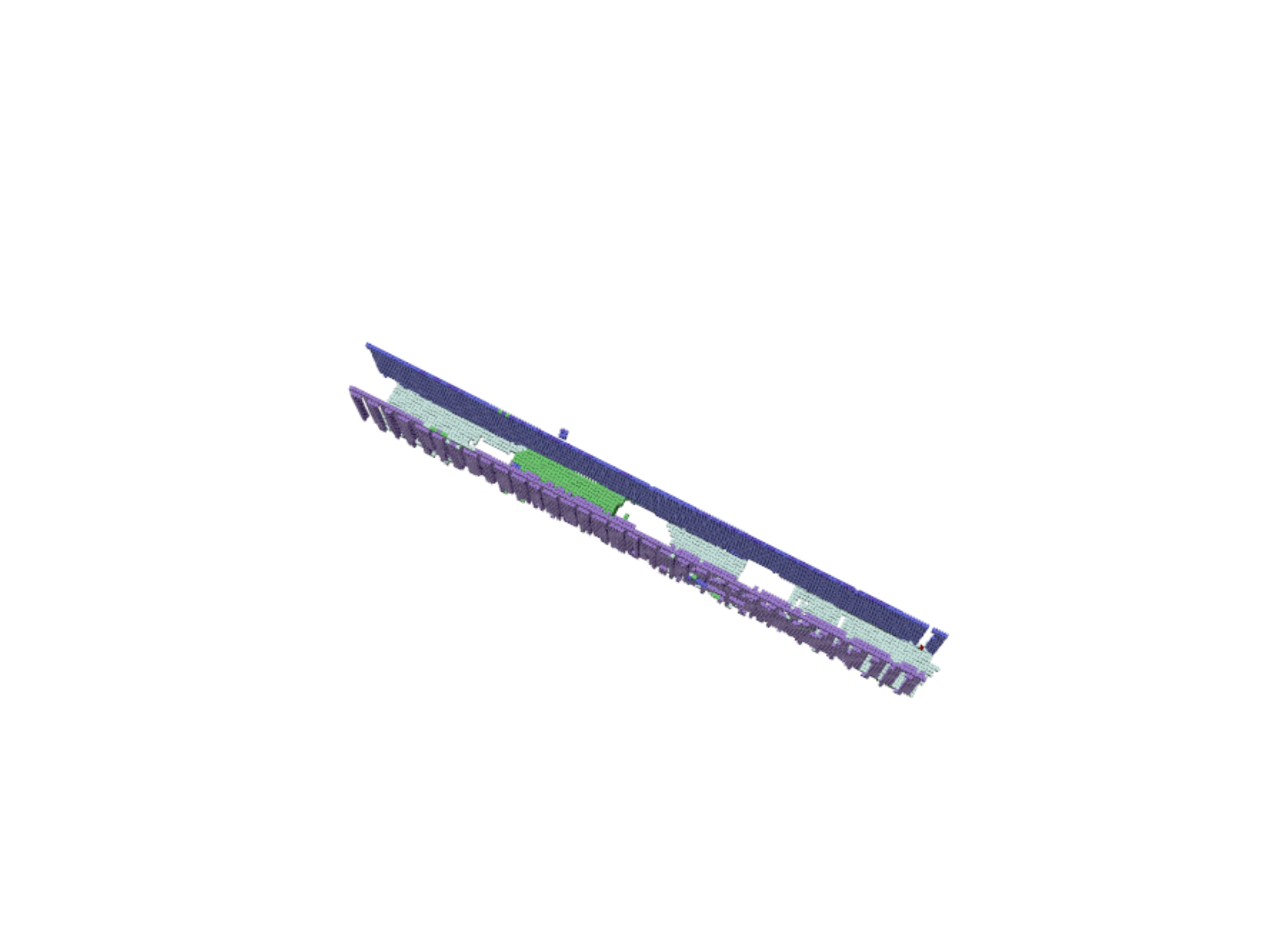}
		\includegraphics[width=0.23\textwidth]{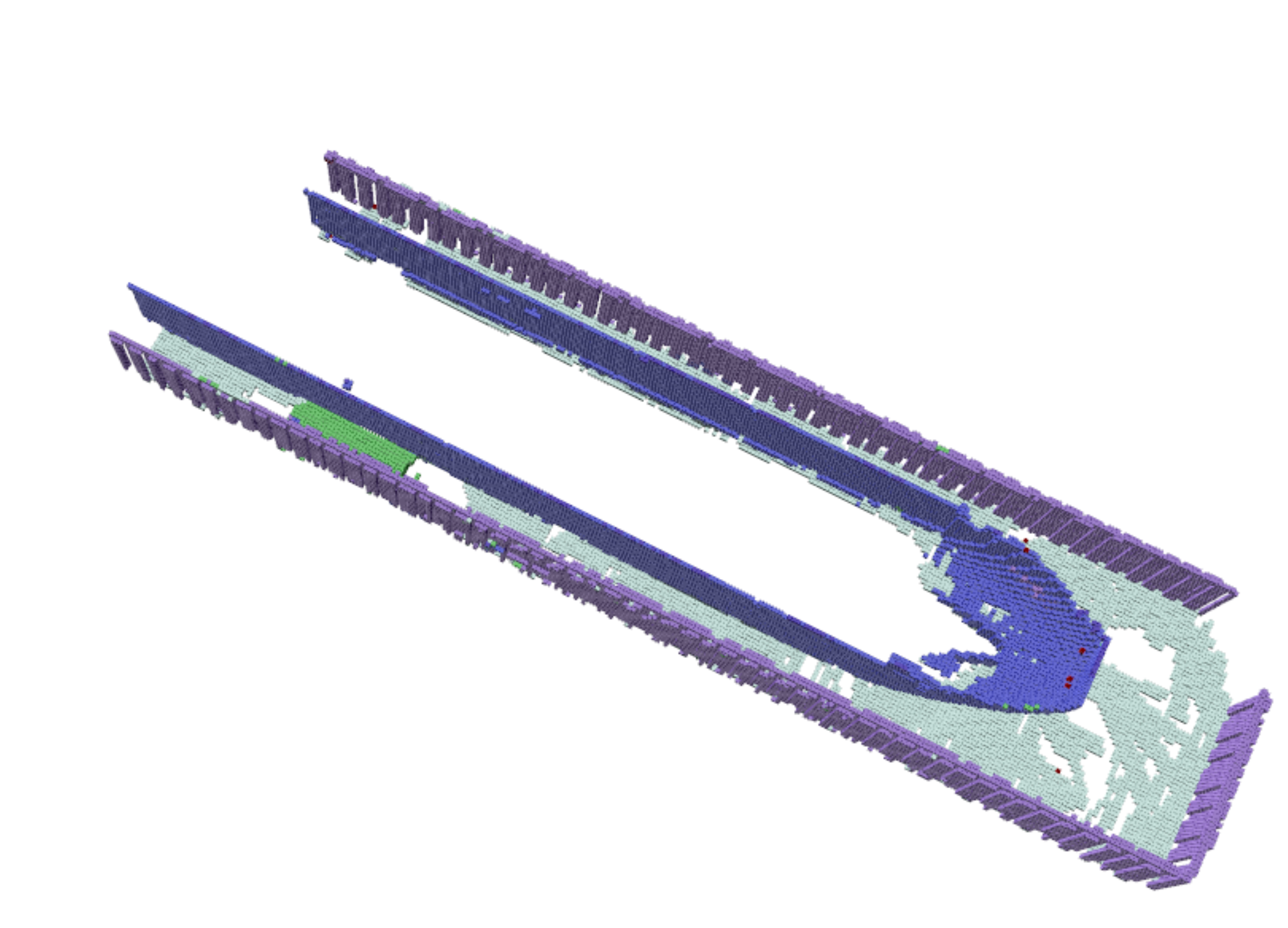}
		\includegraphics[width=0.23\textwidth]{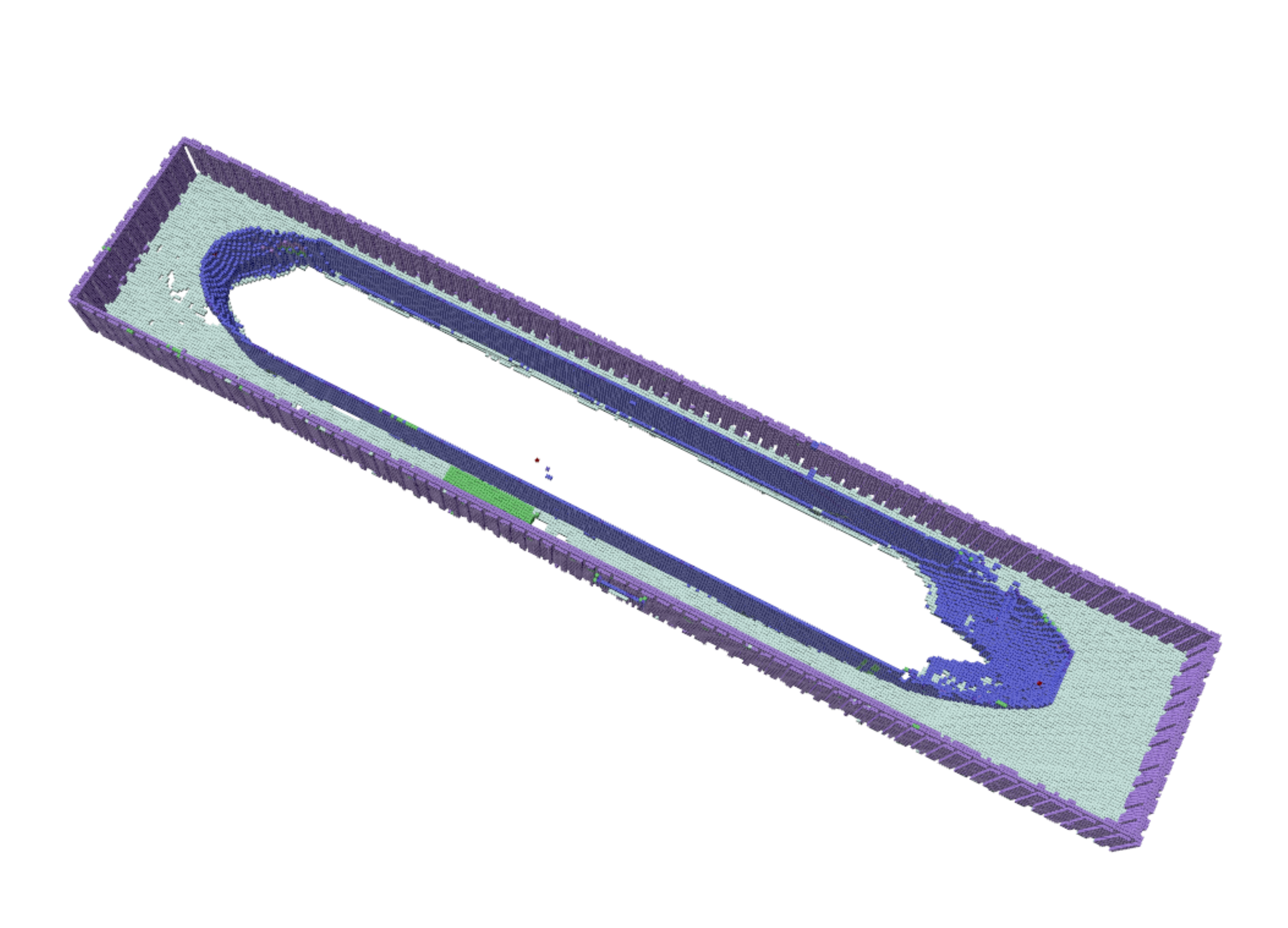}
		\caption{Example reconstruction evolution over \ac{NBV} planning steps with our method. Octomap colored according to most likely semantics.}
	\end{subfigure}
	\begin{subfigure}[b]{0.4\textwidth}
		\centering
		\includegraphics[width=1.0\textwidth]{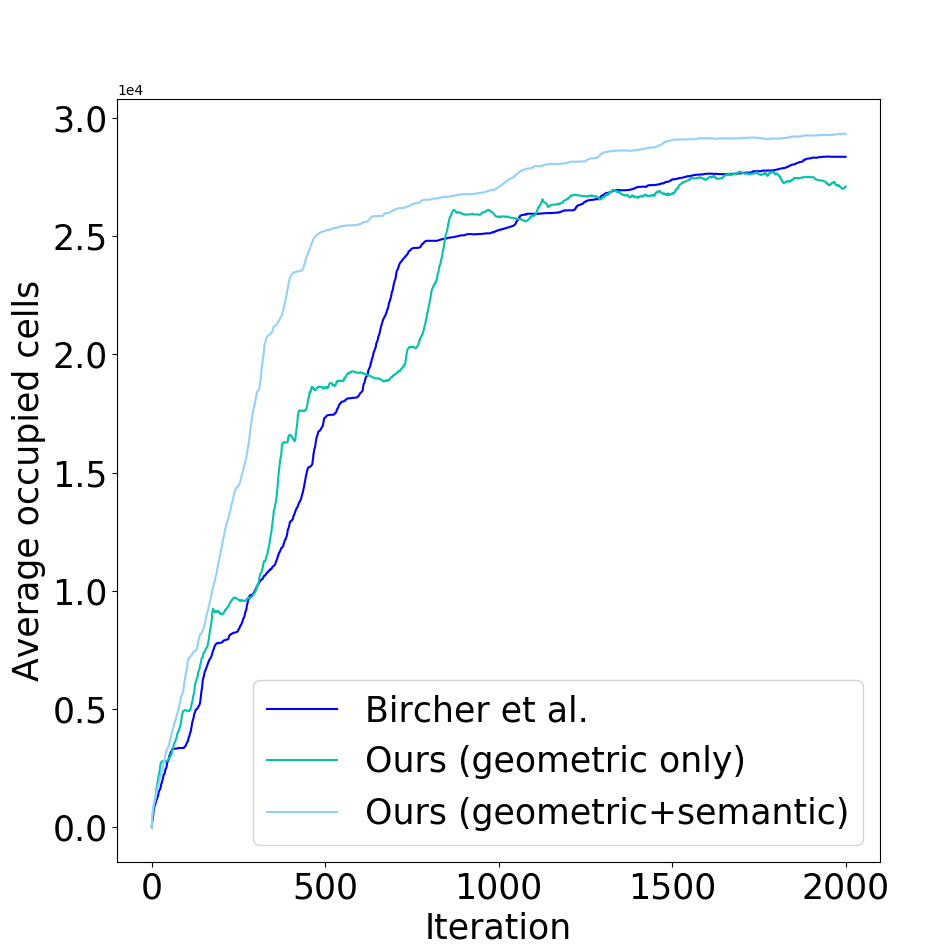}
		\caption{Mapping occupancy.}
	\end{subfigure}
	\begin{subfigure}[b]{0.4\textwidth}
		\includegraphics[width=1.0\textwidth]{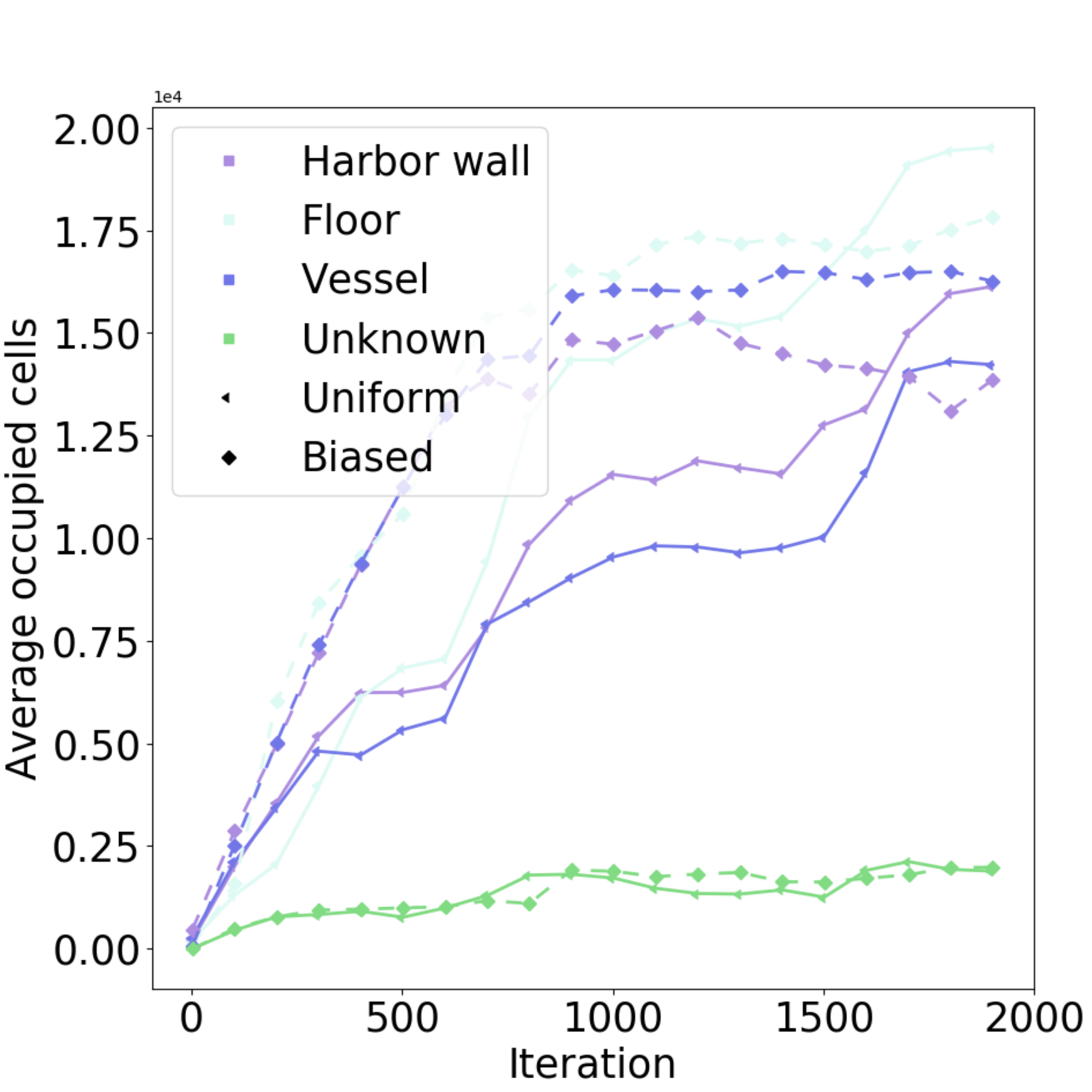}
		\caption{Per-class mapping occupancy.}
    	\label{fig:semantic_simulation_results}
	\end{subfigure}
	\caption{Mapping performance temporal evolution of our \ac{NBV} planning in simulation. Each iteration includes planning, act, and sensing acquisition and fusion in the volumetric-semantic grid.}
	\label{fig:overall_simulation_results}
\end{figure*}

In all our experiments, the axial noise standard deviation scaling factor was set to $\lambda_a=0.005$ and the occupancy probability threshold was set to $P_{occ}=0.7$. In each experiment we let the observer collect $T=2000$ observations (i.e.sense, plan and act iterations). Each experiment was repeated 10 times to average out variability in different simulations. 

\subsubsection{Performance evaluation metrics}
We assess our NBV planning for active and autonomous exploration performance evaluation using the following metrics:
\begin{itemize}
	\item the temporal entropy evolution,
	$\sum_{m_i\in m} H_t(m_i)$
	which is a quality performance measure of the knowledge regarding the surrounding environment, gathered in the probabilistic volumetric map $m$, up to time $t$.
   When  normalized by the number of planning steps it represents the temporal average global information gain per step,
	$\frac{1}{T}\sum_{t=1}^T \sum_{m_i\in m} H_t(m_i)$.
	\item amount of occupied cells (surface coverage)
	$\sum_{m_i\in m} \mathbf{1}\left(P_t(m_i)\right)$
	which is a measure of task-completeness and quantifies the surface covered during reconstruction, where $P_{\text{occ}}$ represents a user specified probability threshold of the volume being occupied.
	When normalized by the number of reconstruction steps it represents the average surface coverage per step,
	$\frac{1}{T}\sum_{t=1}^T \sum_{m_i\in m} \mathbf{1}\left(P_t(m_i)\right)$
\end{itemize}

Finally, we evaluate the computational performance (i.e. efficiency) of the methodologies by measuring sensor fusion and planning times.

\subsubsection{Receding horizon geometric and semantic NBV planning}
The map resolution was set to $\delta=0.4m$ to cope with the task requirements. The assessed map information was considered within the bounds of the motion and planning workspace.
We compare the performance of our method to the state-of-the-art NBV planning approach of~\cite{bircher2016receding}. 
For assessing the advantages of incorporating semantics we considered two distributions for the class weights $C=\{\text{Sky, Floor, Vessel, Harbor Wall, Unknown}\}$. An unbiased uniform one  
\begin{align}
\mathbf{w}^s=\left[0.2,0.2,0.2,0.2,0.2\right]
\end{align}
to impose a purely geometric-driven exploration task, uninformed to semantics, and a biased one to bias exploration towards structures belonging to vessels
\begin{align}
\mathbf{w}^s=\left[0.1,0.1,0.6,0.1,0.1\right]
\end{align}

Table \ref{tab:computational_times} compares the computational performance of the different the baseline mapping \cite{hornung2013octomap} and planning  \cite{bircher2016receding} approaches. 
In this case although planning and mapping runs slower due to the computations involved in estimating the quality (probabilistic) of both geometric and semantic measurements and probabilistic fusion in a volumetric grid, it can still be used at reasonable rates for accurate sensor fusion. Furthermore, as can be seen in table \ref{tab:average_information_gain}, uncertainty aware planning improves mapping quality, due to the fact that previously seen map regions are on average attended more often, until uncertainty becomes negligible.
As can be seen in Table \ref{tab:occupancy_information_semantics} biasing exploration towards specific object classes improves task performance in terms of the mapping occupancy for the class of interest. Furthermore, as can be seen in Fig. \ref{fig:semantic_simulation_results} time-to-full coverage for a specific class (e.g. vessel) of interest is shorter when biasing exploration towards that class (e.g. convergence at around $750$ steps).

\begin{table}
    \caption{Average computational times (ms) per iteration step}
	\begin{center}
		\begin{centering}
			\begin{tabular}{c|cc}
				\toprule
				& Planning & Mapping \\
				\midrule
				Bircher et al. \cite{hornung2013octomap} & 70.3 & 43.3\\
				Ours (geometric only) &82.2&41.2 \\
				Ours (geometric+semantics) &101.5&170.2  \\
			\end{tabular}
		\end{centering}
	\end{center}
	\label{tab:computational_times}
\end{table}

\begin{table}
    \caption{Average information gain per iteration step.}
	\begin{center} 
		\begin{centering}
			\begin{tabular}{c|c}
				\toprule
				& Iterations=2000\\
				\midrule
				Bircher et al. \cite{bircher2016receding}  & 72.19 \\
    			Ours (geometric only) & 82.2 \\
				Ours (geometric+semantics)& 81.08
			\end{tabular}
		\end{centering}
	\end{center}
    \label{tab:average_information_gain}
\end{table}

\begin{table*}[h]
    \vspace{0.2cm}
    \caption{Average occupancy per iteration step.}
    \begin{center} 
		\begin{centering}
			\begin{tabular}{c|cccccc}
			\toprule
& Sky  & Floor & Ship  & Harbor wall & Unknown & Total  \\ 
\midrule
Bircher et al. \cite{bircher2016receding}     & - & - & - & - & - & 52.3\\
Ours (geometric only)                         & - & - & - & - & - & 55.6 \\
Ours (geometric+semantic) (Uniform)           & 0 & 20.2 & 16 & 10 & 4 & 53.2 \\
Ours (geometric+semantic) (Vessel Bias)       & 0 & 10.7 & 35.1 & 4.1 & 5 & 57.4        \\
            \end{tabular}
		\end{centering}
	\end{center}
    \label{tab:occupancy_information_semantics}
\end{table*}

\section{Conclusions}
\label{sec:conclusions}
In this work we have proposed a holistic autonomous navigation framework for \acp{UAV} that incorporates probabilistic semantic-metric mapping representations for receding horizon \acp{NBV} planning. The navigation algorithm leverages both semantic and metric probabilistic gains, in order to decide where to move the UAV, to optimize visual data collection quality of objects of interest, in exploration tasks. 
We have introduced a probabilistic geometric and semantic observation model, and a mapping formulation that allows Bayesian fusion of volumetric and semantic information provided by consumer grade RGB-D sensors and state-of-the-art \acp{DCNN} for 2{D} scene semantic segmentation, in a memory efficient 3{D} octogrid structure. 
Then, we have proposed a real-time, receding-horizon probabilistic path planning approach, that considers both geometric and semantic probabilistic information for planning using \acp{RRT}. Our method is flexible and allows biasing exploration towards specifically known object classes. We have assessed the proposed framework on a realistic simulation environment (Gazebo), and demonstrated the benefits of the pipeline in a set of UAV-based inspection experiments. In the future we intend \added{to evaluate our method in different scenarios, and} extend the proposed approach with the ability to deal with multiple cameras \cite{DBLP:journals/corr/abs-2105-12691}, and schedule sensor acquisition such as to reduce computational load. 

\section*{Acknowledgment}
The authors would like to acknowledge the financial contribution from Smart Industry Program (European Regional Development Fund and Region Midtjylland, grant no.: RFM-17-0020). The authors would further like to thank Upteko Aps for bringing use-case challenges.

\printbibliography

\end{document}